\documentclass[letterpaper, 10 pt, conference]{ieeeconf}

\IEEEoverridecommandlockouts
\usepackage[T1]{fontenc}
\usepackage{float}
\usepackage{textcomp}
\usepackage{amstext}
\usepackage{amssymb}
\usepackage{graphicx}
\usepackage[pagebackref=true,breaklinks=true,letterpaper=true,colorlinks,bookmarks=false]{hyperref}

\hypersetup{pdftitle={Detecting Invasive Insects with Unmanned Aerial Vehicles},
 pdfauthor={Stumph et al.},
 pdfpagelayout=twocolumn, pdfnewwindow=true, pdfstartview=XYZ, plainpages=false}

\makeatletter

\newcommand{\lyxmathsym}[1]{\ifmmode\begingroup\def\b@ld{bold}
  \text{\ifx\math@version\b@ld\bfseries\fi#1}\endgroup\else#1\fi}

\floatstyle{ruled}
\newfloat{algorithm}{tbp}{loa}
\providecommand{\algorithmname}{Algorithm}
\floatname{algorithm}{\protect\algorithmname}

\usepackage[caption=false,font=footnotesize,labelformat=simple]{subfig}
\usepackage{amsfonts}

\usepackage{algorithm}
\usepackage[noend]{algpseudocode}

\IEEEoverridecommandlockouts

\@ifundefined{showcaptionsetup}{}{%
 \PassOptionsToPackage{caption=false}{subfig}}
\usepackage{subfig}
\makeatother

\begin{document}
\title{Detecting Invasive Insects with Unmanned Aerial Vehicles}

\author{{Brian Stumph, Miguel Hernandez Virto, Henry Medeiros, Amy Tabb, Scott
Wolford, Kevin Rice, Tracy Leskey\thanks{Brian Stumph, Miguel Hernandez Virto and Henry Medeiros are with the Department of Electrical and Computer Engineering, Marquette University, Milwaukee, WI, email: brian.stumph@marquette.edu, miguel.hernandez@marquette.edu, henry.medeiros@marquette.edu}
\thanks{Amy Tabb, Scott Wolford and Tracy Leskey are with the USDA-ARS-AFRS, Kearneysville, WV, email: amy.tabb@usda.gov, scott.wolford@usda.gov, tracy.leskey@usda.gov.  Mention of trade names or commercial products in this publication is solely for the purpose of providing specific information and does not imply recommendation or endorsement by the U.S. Department of Agriculture.  USDA is an equal opportunity provider and employer.}
\thanks{Kevin Rice is with the University of Missouri, Columbia, MO, email: ricekev@missouri.edu.}
\thanks{The citation information for this publication is: B. Stumph, M. Hernandez Virto, H. Medeiros, A. Tabb, S. Wolford, K. Rice, T. Leskey, ``Detecting Invasive Insects with Unmanned Aerial Vehicles," in 2019 IEEE International Conference on Robotics and Automation (ICRA), Montreal, QC, Canada, 2019, pp. 648-654.
\href{https://www.doi.org/10.1109/ICRA.2019.8794116}{doi: 10.1109/ICRA.2019.8794116}.}
}}

\maketitle

\begin{abstract}
A key aspect to controlling and reducing the effects invasive insect
species have on agriculture is to obtain knowledge about the migration
patterns of these species. Current state-of-the-art methods of studying
these migration patterns involve a mark-release-recapture technique,
in which insects are released after being marked and researchers attempt
to recapture them later. However, this approach involves a
human researcher manually searching for these insects in large fields
and results in very low recapture rates. In this paper, we propose
an automated system for detecting released insects using an unmanned
aerial vehicle. This system utilizes ultraviolet lighting technology,
digital cameras, and lightweight computer vision algorithms to more
quickly and accurately detect insects compared to the current state of the art. The efficiency and accuracy that this system provides will allow for a more comprehensive understanding of invasive insect species migration patterns. Our experimental results demonstrate that our system
can detect real target insects in field conditions with high precision
and recall rates.
\end{abstract}

\section{Introduction}

Invasive species are often inadvertently transported from their native
environments to new habitats where they tend to have severe negative
impacts on food security, public health, economic interests, and native
species biodiversity \cite{Lockwood_2013_Invasive}. As an example,
in the United States, annual economic losses from invasive species
are estimated at \$120 billion \cite{Pimentel_2005_Losses}, and these
severe impacts are predicted to continue \cite{Paini_2016_Losses}.

Understanding invasive pest migration patterns is a key element to
the mitigation of their damage to both natural environments and agricultural
production. Insect species are highly variable in their dispersal
capacity. For example, invasive brown marmorated stink bugs (BMSB)
are extremely mobile and can fly 115km in 24 hours \cite{Lee_2015_StinkBug,wiman2015factors},
whereas invasive emerald ash borers typically fly less than 100m in
natural habitats \cite{Mercader_2009_EmeraldAshBorer}. Therefore,
the dispersal capacity of newly identified invasive insect species
must be well-defined to determine what strategy (eradication or management)
to initiate.

\begin{figure}[tbh]
\begin{centering}
\includegraphics[width=0.8\columnwidth]{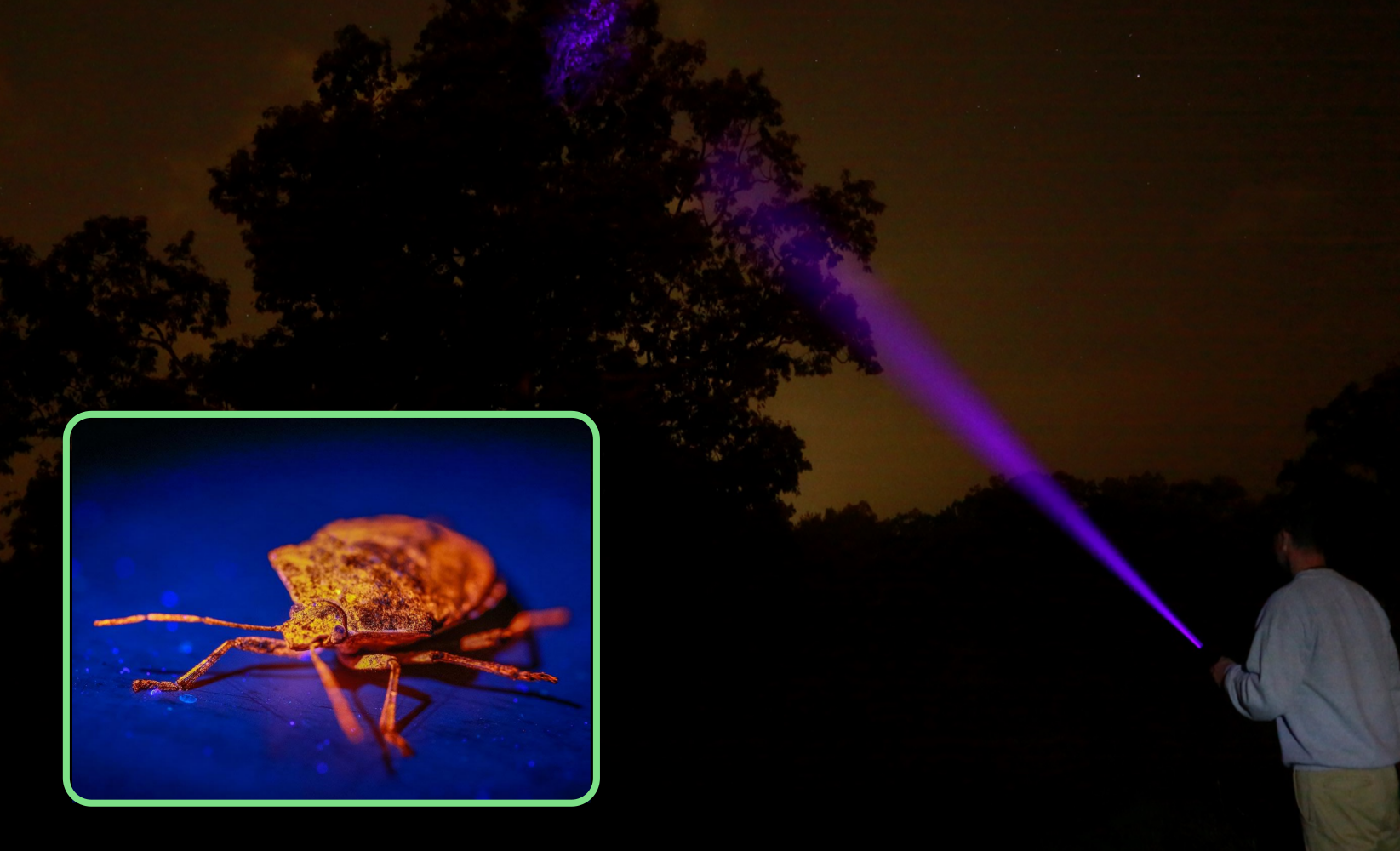}
\par\end{centering}
\caption{Fluorescent-marked BMSB detected in a 30m tree using a hand-held laser
equipped with focusing lens \cite{Rice_2015_Laser}. The inset shows
a fluorescent-marked BMSB illuminated by UV light. \label{fig:Fluorescent-marked-BMSB-detected}}
\end{figure}

Research in this area has focused on the influence of wind and temperature
in long-range migrations, \cite{Taylor_1974_Migration,Drake_1995_Migration},
neglecting the study of short dispersal patterns. Although there are
studies about insect dispersal, they are limited to mark-release-recapture
techniques -- a laborious, time-consuming, and error-prone task.
In addition, because of the labor-intensive nature of the process,
very few samples of the marked insects can be recaptured, typically
< 5\%. This extremely low recapture rate negatively impacts the accuracy
of the resulting dispersal models \cite{Russell_2005_Recapture,Merckx_2009_Recapture,nazni2005determination}.

In this paper, we describe a novel aerial system that attempts to
solve the limitations of the manual methods described above. The proposed
approach uses an Unmanned Aerial Vehicle (UAV) to scan a region of
interest and detect target insects using a recently developed system
of ultraviolet (UV) lights. With the UAV equipped with a UV light
source and video camera, we record an aerial video of the area. We
later process the video offline to obtain an accurate count of
the marked insects released in the field. 

The video processing pipeline described in this paper employs multi-level
color channel thresholding to segment the marked insects from the
background. It consists of three main steps. First, we extract the
region of interest (ROI) from the video frames, keeping only the region
surrounding the area illuminated by the UV light. Then, we identify the
illuminated insects using color channel thresholding in the RGB and
HSV color spaces. Finally, we use watershed segmentation \cite{Sun_watershed}
to separate nearby insects that might have been identified as a single
insect in the previous step. Our experiments demonstrate the detection
of marked insects in field conditions.

The contributions and impact of this work are as follows: 1) We propose
a novel data acquisition system based on a UAV that allows us to quickly
search for insects in a large area of interest, 2) we design a low
complexity computer vision algorithm to robustly detect and count
the insects observed by our system, and 3) we evaluate the data acquisition
system in open field situations using industry standard benchmarks.
The impact of this work represents a considerably faster, economical,
and accurate alternative to current manual mark-release-recapture
methods for insect pest monitoring. Moreover, as there is no need
to recapture the insects after they are detected, our non-intrusive
method allows for researchers to dynamically localize insects over
time.

\section{Related Work}

For over a century, entomologists have used mark-release-recapture
techniques to track insect movement. These initial efforts traditionally
used paint and dyes as markers \cite{Geiger_1919_Marker}. In recent
years, protein markers were developed to quickly mark thousands of
individual insects within minutes \cite{Hagler&Durand_1994_Marker,Hagler&Jackson_2001_Marker},
enabling more sophisticated studies. However, protein marking involves
inherent challenges. First, the proteins degrade quickly under field
conditions. In addition, a time-consuming process based on enzyme-linked
immunosorbent assays (ELISA) is needed to determine whether a specific
protein marker is present \cite{Buss_1997_ELISA}. Finally, because
of the ELISA test itself, the target insect is ultimately destroyed
thereby eliminating the possibility of tracking the same insect more
than once.

\subsection{Ultraviolet light tracking}

The use of methods based on fluorescent pigment is a simple alternative
that can effectively mark millions of insects in a short amount of
time \cite{Stern&Muller_1968_Fluorescent,Schroder&Mitchell_1981_Fluorescent}.
Fluorescent markings (see Figure \ref{fig:Fluorescent-marked-BMSB-detected})
glow under UV light by transforming radiant energy from the UV band
to longer wavelengths that are detectable by the human eye. However,
detection distances of fluorescent powders with UV light sources are
significantly short -- less than a few meters -- even in dark conditions.
Thus, many studies employing fluorescent pigment marking require recaptured
organisms to be taken to the laboratory to verify marks under UV lamps
\cite{Longland&Clements_1995_UVLamps,Narisu_1999_UVLamps}, rather
than in the insect's initial location. 

Rice et al. described in \cite{Rice_2015_Laser} a novel method for
detecting fluorescent-marked insects. They use hand-held UV lasers
with focusable lenses which allow the beam to be widened significantly
for a larger search area and increased detection distance up to 40m. The method also enables non-destructive scanning from the ground
of previously inaccessible habitats, including tree canopies and aquatic
habitats. These contributions have resulted in increased recovery
rates and the amount of information gathered. Additionally, utilizing
UAVs in conjunction with UV lights has been proposed in other research
applications. One such example is using a UAV to inspect transmission
lines, while implementing a UV light to highlight problems with high
voltage coronas \cite{Rymer_UV_detection}. This demonstrates the
advantage of using UV light to assist with UAV vision operations.

\subsection{Agricultural applications of unmanned aerial vehicles}

Recently, there has been increasing interest in the use of UAVs in
agricultural research, with predictions that it will revolutionize
spatial ecology \cite{Anderson&Gaston_2013_UAV}. UAVs provide the
ability to collect remote data at an unprecedented scale and sampling
rates at a fraction of the cost of previous methods such as satellites
or manned aircraft \cite{Zhang_2012_application}. Current examples
of research applications include methods of classification and monitoring,
such as weed classification \cite{Lottes2017UAVbasedCA} and applying
a multispectral imaging system for crop monitoring \cite{Oca_Crop_Monitoring}.
Similar to our proposed system of mapping insects in a field, other
related research topics include aerial mapping of rice crops \cite{Rojas_Rice_crops}.
Similar agricultural mapping techniques involve using swarms
of UAVs to map weeds in large fields \cite{Albani2017FieldCA} and
terrain surveying of disjointed fields aided by UAVs and path planning
\cite{Vasquez_Terrain_Surveying}. Lastly, UAVs can provide a system
for detection and estimation, such as using the vehicles to provide
vital data on plant stress \cite{Bhandari_Precision_Agriculture}.

The UAV sensing modality is determined by the application needs as
well as the payload capacity of the aircraft. Hyperspectral and multispectral
cameras are among the most popular sensors, but regular and thermal
cameras, lidar and radar imaging and even chemical sensing have been
applied in specialized areas \cite{Pajares_2015_Sensors}. Recently,
radio tags have been employed to track invasive fish species using
UAVs \cite{Plonski2017EnvironmentEI}. However, no applications of
UAVs to track insect movement and dispersal in field conditions have
been proposed so far.

\section{Insect detection system}

In this section, we describe the hardware and software components
of the proposed detection system. We designed the system with the
assumption that insects have been previously coated with a fluorescent
pigment. It is also assumed the data acquisition process is carried
out at night with limited artificial illumination sources other than
the UV light system. The insect detection process consists of two
primary components, the aerial data acquisition and the insect detection
and segmentation algorithm. Data acquisition consists of flying the
UAV over the area under study with the camera and UV light source facing down, illuminating and filming any coated insects on the ground. Once data
acquisition is completed, the video files are downloaded onto a workstation
and used as input to the software pipeline. This pipeline utilizes
multi-channel thresholding techniques to detect all the insects in
the video.

\begin{figure}[tbh]
\begin{centering}
\subfloat[]{\includegraphics[width=0.48\columnwidth]{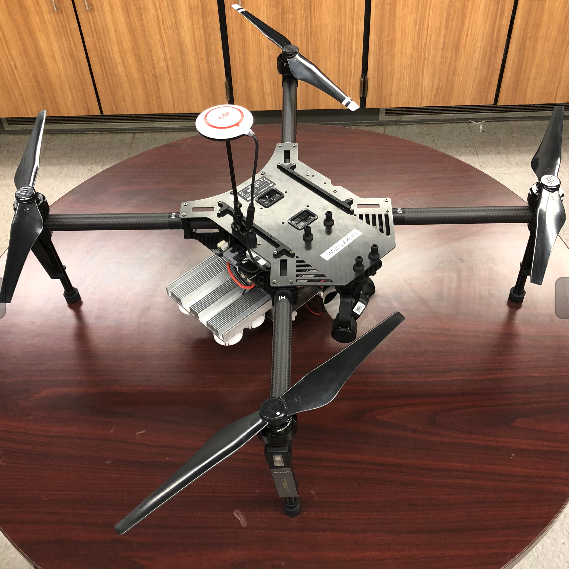}

}~\subfloat[]{\includegraphics[width=0.48\columnwidth]{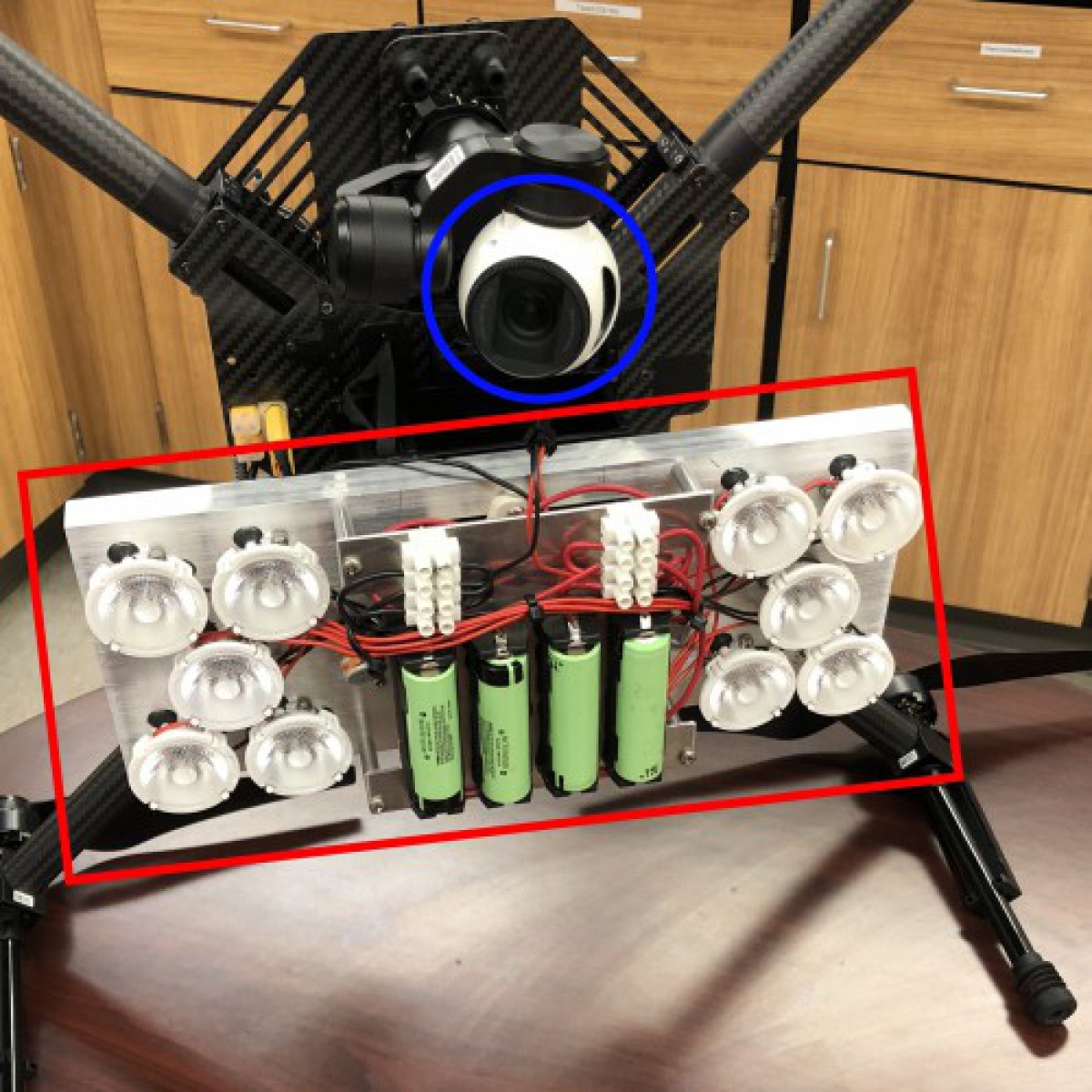}

}
\par\end{centering}
\caption{a) Matrice 100, equipped with a Zenmuse X3 camera and gimbal unit and a UV light system. b) Close-up view of the bottom of the UAV. The blue circle denotes the camera and the red rectangle indicates the UV LED array. \label{fig:FlameWheel-F550,-equipped}}
\end{figure}

\subsection{Hardware components}

Our system consists of a UAV equipped with a set of ten ultraviolet
lights and a high-resolution camera and gimbal unit. The UAV is a
Matrice 100 model -- a four-propeller UAV that is ready-to-fly off
the shelf yet still fully customizable (see Figure \ref{fig:FlameWheel-F550,-equipped}).
A benefit of the Matrice 100 is its extended flight time of up to
40 minutes. This is essential for the plausibility of our proposed
system, as large fields will take considerable time for the UAV to
cover. 

The camera and gimbal unit is the Zenmuse X3. The Zenmuse X3 is an
all-in-one gimbal and camera tool that can produce videos with $4096\times2160$
pixel resolution and 3 axis movement stability. The gimbal is set
to be pointed perpendicular to the ground at all times. The 3-axis
stability allows for clear, non-blurry footage to be collected while
the UAV is moving. The camera's parameters can be adjusted through
the DJI Ground Station Pro app for iOS devices. All conducted tests
would be in a low-light environment, so we used an ISO rating of 1600
and a shutter speed of 1/25 second for all video footage.

The UV illumination system is attached to the bottom frame of the
UAV and can be controlled by a remote transmitter. The UV lights consist
of high-power violet LEDs with a wavelength of 395 nm. Each LED is
encapsulated by a narrowing lens that focuses the light emitted by the LED. The lights are attached to a $12.7\times30.5$ cm aluminum
heatsink (see Figure \ref{fig:FlameWheel-F550,-equipped}) and a set
of four 3.7V Li-ion batteries with 3400mAh capacity to supply power
to the UV lights. The power source is controlled with an RF remote control
relay switch fastened between the heatsink and battery pack. The light
system is secured to the UAV using four fastening screws so that the
UV lights are pointed downward. 

Although the camera is mounted on a gimbal, the illumination system
is not. This inevitably causes the UV projection to always be perpendicular
to the UAV rather than the ground when the UAV is in motion. To minimize
the movement of the UV-illuminated area, we installed the UV LED array in
the center of the vehicle's body frame, which corresponds to the pivot
point of the pitch, roll, and yaw motions. This layout also improves
flight dynamics and stability, as it reduces the moment of inertia
caused by the weight of the system (1029g with batteries). Additionally,
the UAV is flown at the relatively slow speed of 1 m/s to reduce the angular tilt caused by the forward movement of the UAV.

\begin{figure*}[t]
\subfloat[]{\includegraphics[width=0.162\paperwidth]{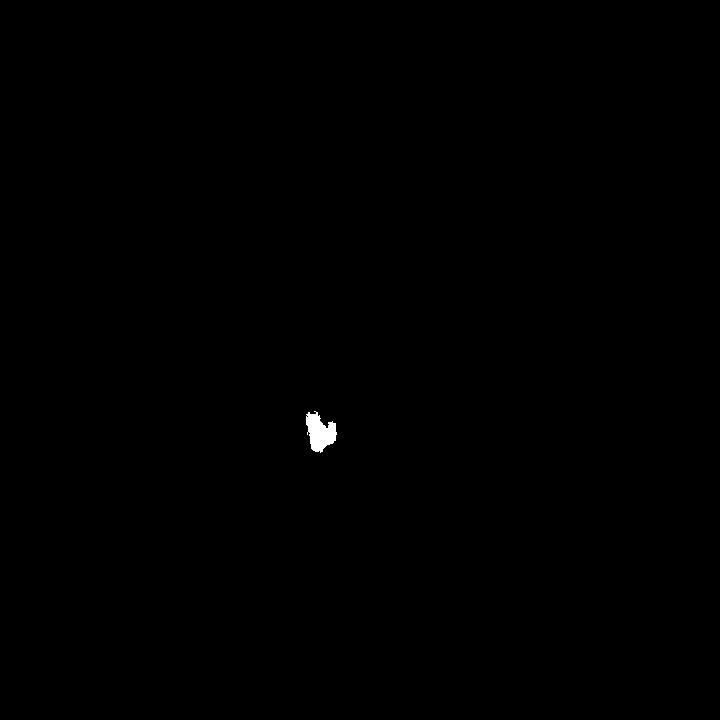}

}\subfloat[]{\includegraphics[width=0.162\paperwidth]{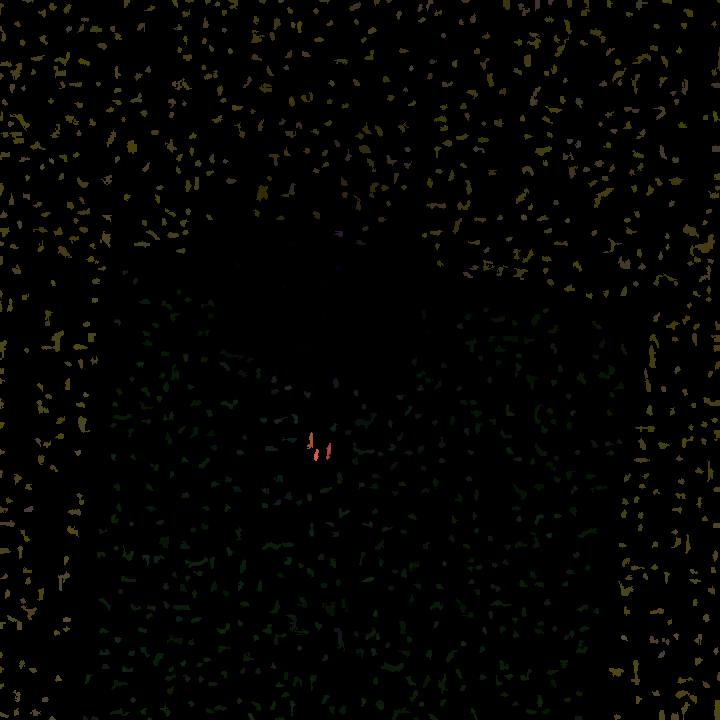}

}\subfloat[]{\includegraphics[width=0.162\paperwidth]{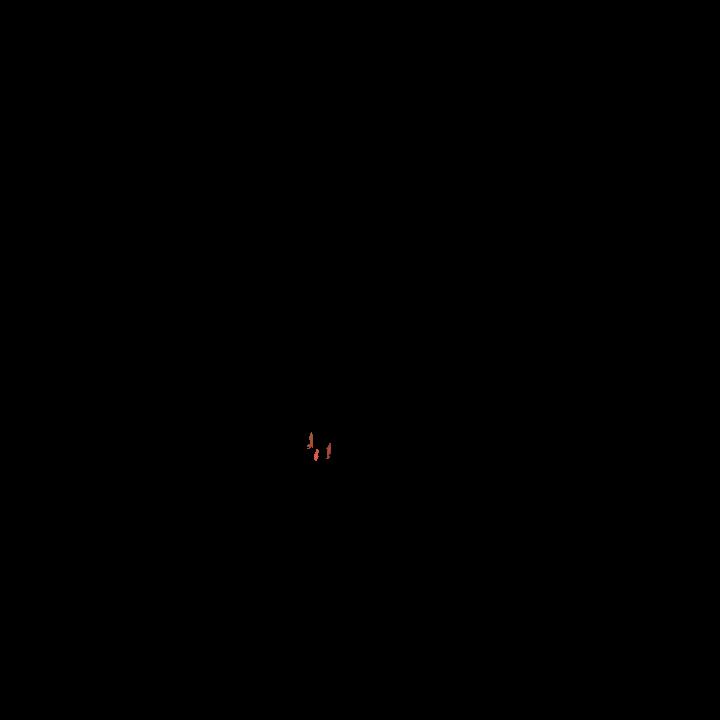}

}\subfloat[]{\includegraphics[width=0.162\paperwidth]{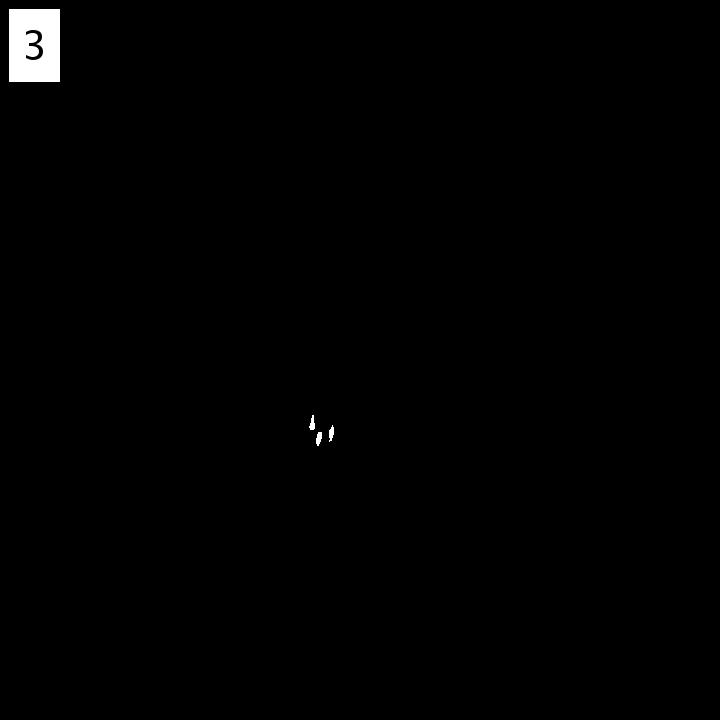}

}\subfloat[]{\includegraphics[width=0.162\paperwidth]{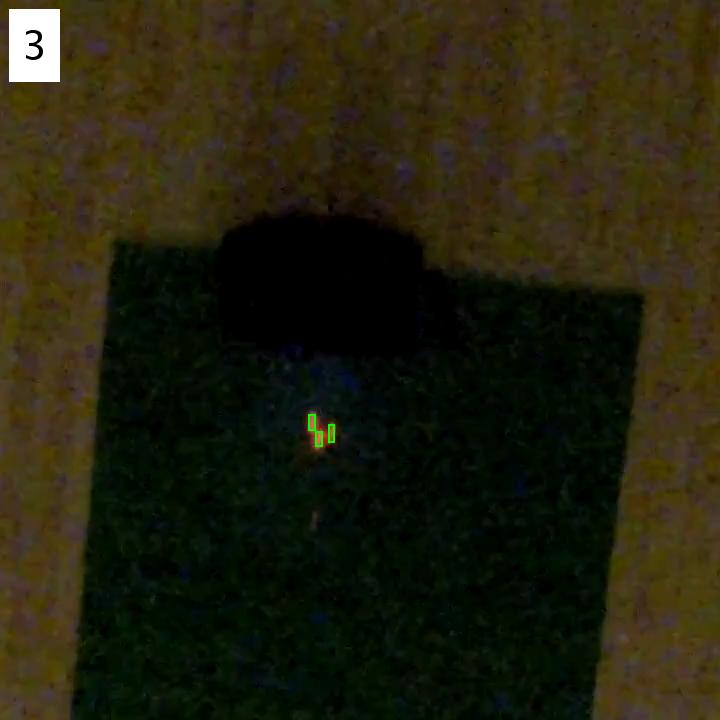}

}

\caption{Illustrative intermediate steps of the proposed software pipeline.
a) Original image $\mathbb{I}_{i}^{(r)}$ multiplied by thresholded
image $\mathbb{I}_{i}^{(d)}$. b) $\mathbb{I}_{i}^{(w)}$ regional
intensity peaks of $\mathbb{I}_{i}^{(r)}$. c) $\mathbb{I}_{i}^{(s)}$,
individual insect detections, obtained by applying $\mathbb{I}_{i}^{(d)}$
as a mask to $\mathbb{I}_{i}^{(w)}$. d) Binarized $\mathbb{I}_{i}^{(s)}$
with the blob counter added on the top left of the image. e) Original
image frame $\mathbb{I}_{i}^{(r)}$ with counter and overlaid bounding
boxes around the detected insects.\label{fig:5pipeline_steps}}
\end{figure*}

\subsection{Software pipeline\label{subsec:Software-pipeline}}

The video recorded during the flight, $\ensuremath{\mathbb{V}}$,
consists of a sequence of $n$ frames
\begin{equation}
\ensuremath{\mathbb{V}}=\left\{ \mathbb{I}_{1},\mathbb{I}_{2},...,\mathbb{I}_{n}\right\} ,\label{eq:V}
\end{equation}
where each frame $\mathbb{I}_{i}\in\mathbb{R}^{h\times w\times c}$,
with $h$ and $w$ representing the height and width of the image,
and $c$ its number of color channels. Thus, each image pixel $p_{x,y}^{i}$
is a $c$-dimensional vector, where $x$ and $y$ are its coordinates
in the image. The output video $\ensuremath{\mathbb{V}}$ is then
used as the input to Algorithm \ref{alg:VideoProcessing}, which summarizes
the processing steps carried out after each data collection flight.

\begin{algorithm}[tbh]
\begin{algorithmic}[1]
\Require{UAV camera video file $\mathbb{V}$.}
\Ensure{Video $\mathbb{V'}$ showing the detections and detection counts per frame.}
\For {Each frame $\mathbb{I}_i \in \mathbb{V}$}
\State {Extract an $h'\times w'$ region of interest $\mathbb{I}^{(r)}_i$ from   $\mathbb{I}_i$.}
\State {Create the detection image $\mathbb{I}^{(d)}_i$ by thresholding $\mathbb{I}^{(r)}_i$ \par in the HSV and RGB color spaces.}
\State {Combine regional hue maxima of $\mathbb{I}^{(r)}_i$ with $\mathbb{I}^{(d)}_i$ to \par split groups of insects into individual detections, \par $d_i$, creating  $\mathbb{I}^{(s)}_i$.}
\State {Count and locate $d_i$ by analyzing the blobs of the \par binarized $\mathbb{I}^{(s)}_i$. }
\EndFor
\end{algorithmic}

\caption{\label{alg:VideoProcessing}Insect detection video processing algorithm.}
\end{algorithm}

\textit{1) Step 1: Region of interest selection:} To reduce the execution
time of our software pipeline, a region of interest is selected in
relation to the middle of each frame. A supplemental advantage of
selecting a region of interest is reducing the amount of noise on
the edges of the frame since all the insects should be visible only within
the ultraviolet light-illuminated portion of the frame. The field of view (FOV) of the light system ($30{^\circ}$) is significantly narrower than the FOV of the Zenmuse X3 ($94{^\circ}$). Therefore, we determined the size of the image $\mathbb{I}_{i}^{(r)}$, $h'\times w'$, experimentally to ensure
the UV light projection always remains within the region of interest.
The determined values were $720$ x 720 pixels, which corresponds
to 25\% of the original image size. This allows for a relative angle
of up to 16$\lyxmathsym{\textdegree}$ between the camera and the
light beam. 

\textit{2) Step 2: Thresholding:} To identify the insects in $\mathbb{V}$,
we segment the pixels of $\mathbb{I}_{i}^{(r)}$ as belonging to the
foreground (insects) or background using simple color thresholding.
This is plausible because of the distinctive color of the marked insects
with respect to the background. The result of this step is $\mathbb{I}_{i}^{(d)}$,
an image displaying the detected insects over a black background,
as shown in Figure \ref{fig:5pipeline_steps}(a). To achieve this
result, we used two different color space representations, RGB (red,
green, blue) and HSV (hue, saturation, value).

The fluorescent powder used shows a pink color when illuminated with
UV light. Pink exhibits very low green values compared with its red
and blue components. We enforce this relation between the three color channels in our insect thresholding algorithm. That is, let $p=\left[p_{r},p_{g},p_{b}\right]$
be a pixel in $\mathbb{I}_{i}^{(r)}$, then $p$ is considered a foreground
pixel as long as $p_{r}>p_{b}>p_{g}$.

RGB thresholding alone, however, is not robust enough against reflective
surfaces such as dewy grass. In our case, the reflection of the UV
light (violet color) on some surfaces can result in dark pink pixels,
which can be mistaken for insects. To improve the robustness of our
method, we incorporate HSV color space thresholding. We use a set
of nine calibration images of insects illuminated by the UV light
on a non-reflective background to determine $\mu_{v}$, the average
values of the brightness component of the reflected light. We then
use this value as a lower detection threshold. That is, let $p=\left[p_{h},p_{s},p_{v}\right]$ be a pixel in $\mathbb{I}_{i}^{(h)}$, the HSV representation of $\mathbb{I}_{i}^{(r)}$,
then for $p$ to be considered a foreground pixel it has to satisfy
$p_{v}>\mu_{v}$. The determined value $\mu_{v}$ was 40. Additionally,
the value channel threshold allows for further differentiation from
the black background, as black pixels have very low brightness values.
We have determined experimentally that our method is not sensitive
to the number of images used for calibration. The method shows identical
results if the number of calibration images is between five and eighteen.

The bright red and pink color of the marked insects correspond to
both very high and very low values in the hue channel, leading us
to choose an upper and lower threshold requirement for hue levels.
The values are denoted by $h_{ut}$ and $h_{lt}$ for the upper and
lower hue thresholds, respectively. Inversely, we found that marked
insects typically exhibited saturation values in the middle of the
saturation spectrum. Thus, we again use upper and lower thresholds
for the saturation channel, denoted by $s_{ut}$ and $s_{lt}$, respectively.
That is, for a pixel $p$ to be considered a foreground pixel, it must satisfy $p_{h}>h_{ut},$ $p_{h}<h_{lt}$, $s_{lt}<p_{s}<s_{ut}$. The following optimal threshold values were found throughout our experiments: $h_{ut}$ = 220, $h_{lt}$ = 25, $s_{lt}$ = 90, $s_{ut}$ = 255.

\textit{3) Step 3: Segmentation:} Thresholding alone is not sufficient
to identify multiple insects in close proximity to each other. Groups
of coated insects represent very brightly colored areas, and the gap
between them is often detected as part of the foreground (Figure \ref{fig:5pipeline_steps}
(a)). To overcome this difficulty, we use watershed segmentation \cite{Parvati_2009_Segmentation}
to split groups of insects into individual detections. By applying
watershed segmentation to $\mathbb{I}_{i}^{(r)}$, we generate an
image of local intensity peaks, $\mathbb{I}_{i}^{(w)}$ (Figure \ref{fig:5pipeline_steps}
(b)). Then we use $\mathbb{I}_{i}^{(d)}$ as a mask for $\mathbb{I}_{i}^{(w)}$
so that only the peaks that are inside the thresholded blobs of $\mathbb{I}_{i}^{(d)}$
are maintained. The new image $\mathbb{I}_{i}^{(s)}$ is given by
\begin{equation}
\mathbb{I}_{i}^{(s)}=\mathbb{I}_{i}^{(d)}\odot\mathbb{I}_{i}^{(w)}\label{eq:I_i}
\end{equation}
where $\odot$ represents elementwise multiplication. As Figure \ref{fig:5pipeline_steps}
(c) illustrates, all detected insects can be clearly identified in
the resulting image $\mathbb{I}_{i}^{(s)}$. 

\textit{4) Step 4: Counting:} To count the detected insects $d_{i}$
in $\mathbb{I}_{i}^{(s)}$, we first binarize the image using the
thresholds and segmentation steps from steps two and three. We subsequently
perform blob analysis using MATLAB's Computer Vision Toolbox\footnote{https://www.mathworks.com/products/computer-vision.html},
thereby counting the number of blobs (groups of foreground pixels)
in the image. This step additionally allows us to apply a detection
size filter to discard blobs that contain fewer than a certain number
of pixels $n_{p}$.

To calculate the total number of insects in the video, we create a
global counter that is incremented whenever the number of blobs in
$\mathbb{I}_{i}^{(s)}$ is higher than the number of blobs in $\mathbb{I}_{i-1}^{(s)}$.
Figure \ref{fig:5pipeline_steps} (d) shows the result of this step.
Note that this method does not take into consideration potential double-counting
caused by scanning an image area multiple times. Addressing this limitation
is part of our future work.

Finally, for visualization purposes, we highlight each blob with a
green bounding box, as show in Figure \ref{fig:5pipeline_steps} (e).

\section{Experimental Results}

In this section, we evaluate the performance of the proposed system
in our testing location, a grassy outdoor field. We compare our method
with a baseline approach that uses Otsu's algorithm \cite{otsu1979threshold}
to determine the optimal luminance threshold for the video. 

\subsection{Baseline method}

\begin{figure}[tbh]
\begin{centering}
\subfloat[\label{fig:Otsu_a}]{\begin{centering}
\begin{minipage}[t]{0.27\columnwidth}%
\begin{center}
\includegraphics[width=1\columnwidth]{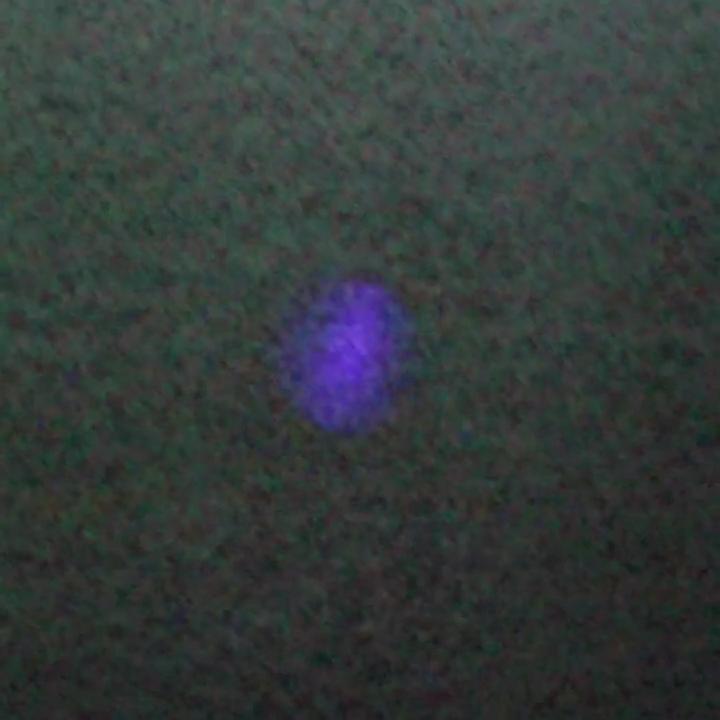}
\par\end{center}
\begin{center}
\includegraphics[width=1\columnwidth]{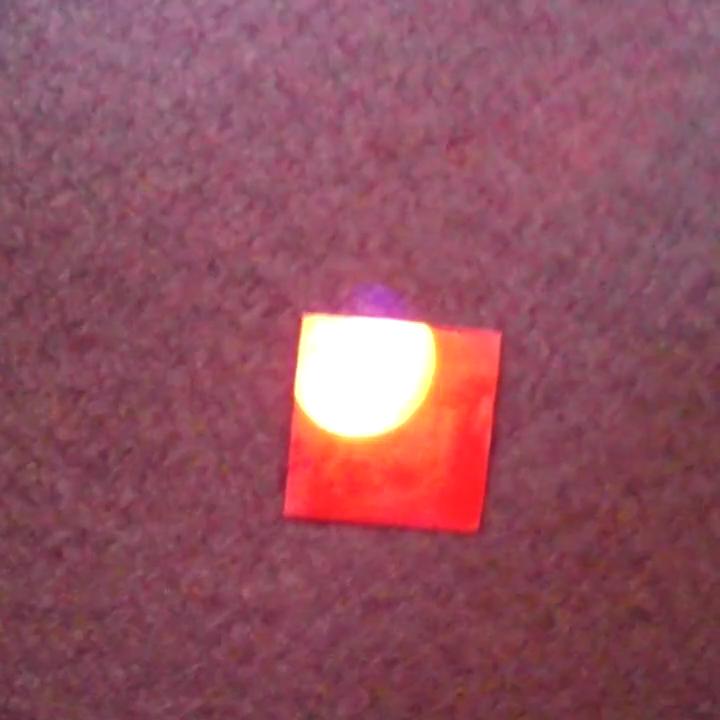}
\par\end{center}%
\end{minipage}
\par\end{centering}
}\,\subfloat[\label{fig:Otsu_b}]{\centering{}%
\begin{minipage}[t]{0.27\columnwidth}%
\begin{center}
\includegraphics[width=1\columnwidth]{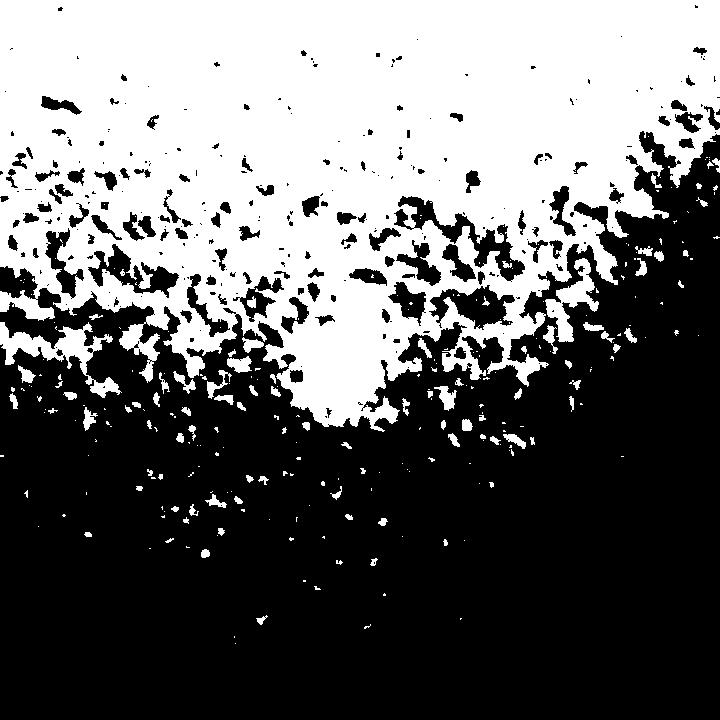}
\par\end{center}
\begin{center}
\includegraphics[width=1\columnwidth]{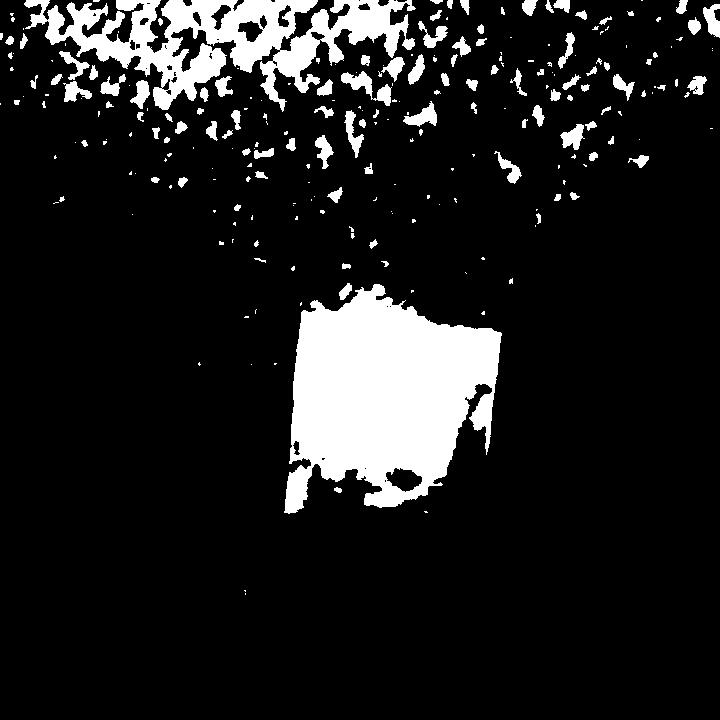}
\par\end{center}%
\end{minipage}}\enskip{}\subfloat[\label{fig:Otsu_c}]{\begin{centering}
\begin{minipage}[t]{0.27\columnwidth}%
\begin{center}
\includegraphics[width=1\columnwidth]{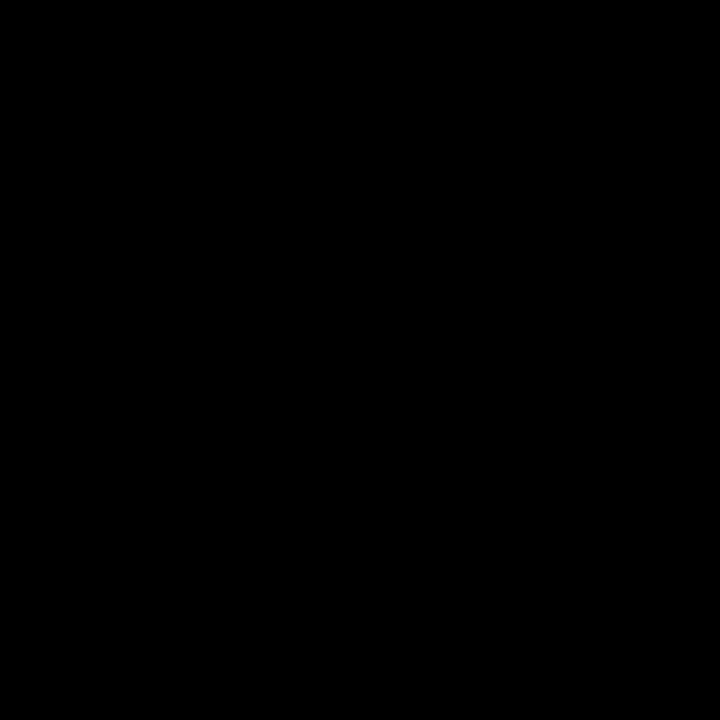}
\par\end{center}
\begin{center}
\includegraphics[width=1\columnwidth]{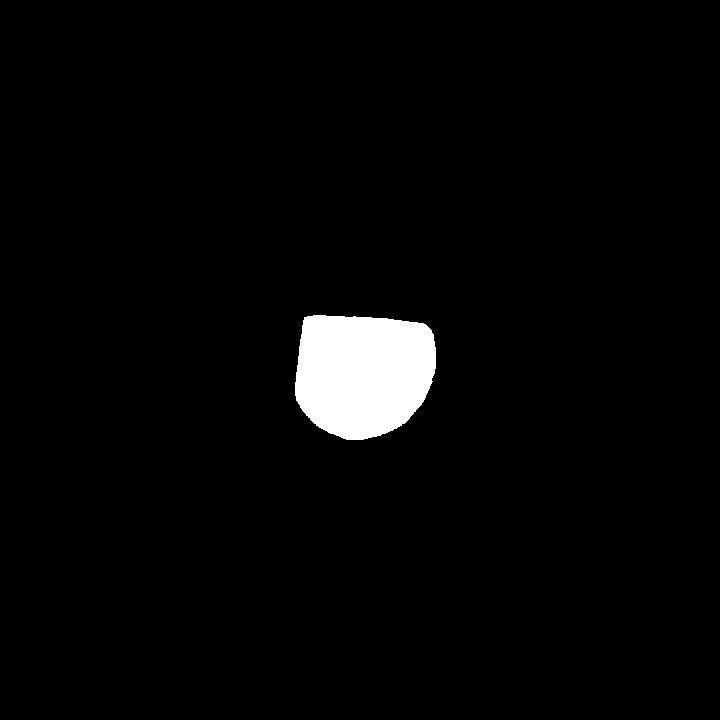}
\par\end{center}%
\end{minipage}
\par\end{centering}
}
\par\end{centering}
\caption{Binarization of two video frames using Otsu's algorithm.
a) Original frames, one without a fluorescent object and the other
with an illuminated coated board. b) Binarization using different
thresholds for each frame ($t_{o,i}=59$ without fluorescent object
and $t_{o,i}=119$ with the object), c) Binarization of the frames
using the maximum threshold value of the video according to Eq. \ref{eq:threshold}
(in this case, $t_{m}=176$).}
\end{figure}

To the best of our knowledge, no automated methods to segment fluorescent
insects currently exist. Therefore, we select Otsu's algorithm as
a baseline approach for comparison. Figures \ref{fig:Otsu_a} and
\ref{fig:Otsu_b} illustrate that when no fluorescent target is illuminated,
Otsu's threshold generates significant amounts of clutter. To address
this problem, we search the entire video for the maximum threshold
value $t_{m}$ obtained applying Otsu's algorithm to each frame and
use that value for all the video frames on a subsequent pass, i.e.,
\begin{equation}
t_{m}=\max_{\mathbb{I}_{i}\in\mathbb{V}}t_{o,i},\label{eq:threshold}
\end{equation}
where $t_{o,i}$ is Otsu's threshold for frame $\mathbb{I}_{i}$.
Figure \ref{fig:Otsu_c} shows the results of applying this method
to the two images shown in Figure \ref{fig:Otsu_a}.

\subsection{Target species and marking procedure}

\begin{figure}[tbh]
\centering{}\subfloat[\label{fig: test-bugs}]{\includegraphics[clip,width=0.48\columnwidth]{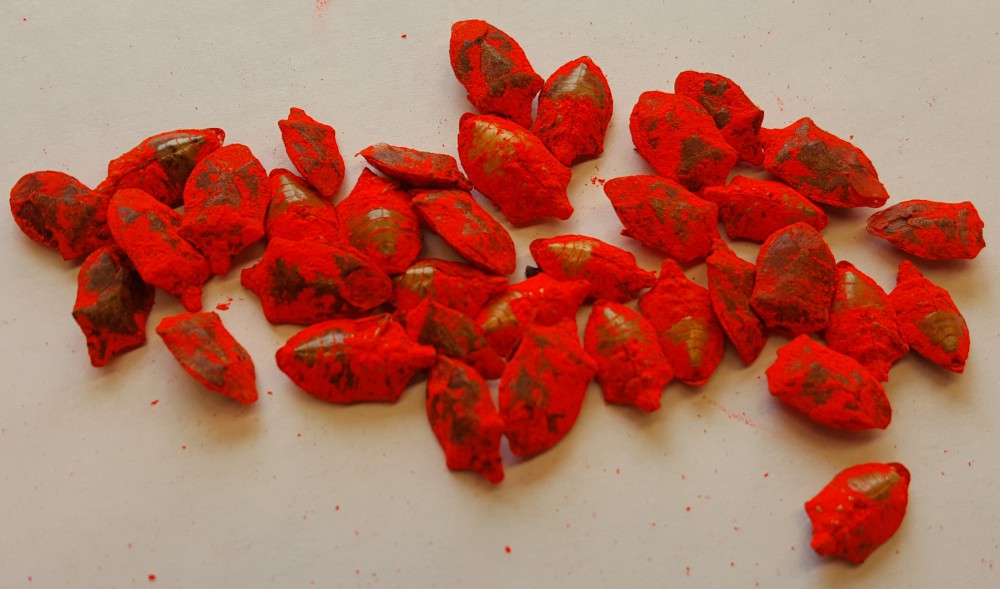}}\,\subfloat[\label{fig: test-bugs-ruler}]{\includegraphics[clip,width=0.48\columnwidth]{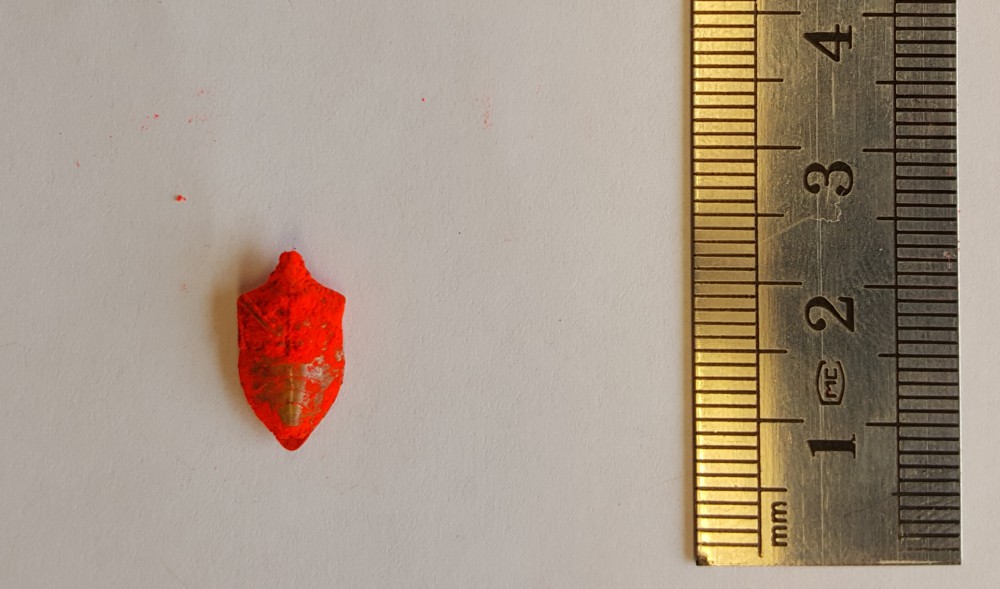}}\caption{a) Insects used during the design and evaluation of our system. b)
Single insect with ruler for reference.}
\end{figure}

All the experiments described in this section were carried out using real
fluorescent-marked insects. In particular, we used 36 fluorescent-marked
brown marmorated stink bugs (BMSB) with sizes between 13.5 x 7 mm
to 16 x 8 mm (Figure \ref{fig: test-bugs}). To coat the insects, we individually placed them in a plastic cylinder with 2 g of red fluorescent powder (BioQuip, Rancho Dominguez, CA) and gently shook it for five seconds.

\subsection{Test site}

\begin{figure}[tbh]
\centering{}\subfloat[]{\includegraphics[viewport=0bp 0bp 595bp 595bp,width=0.47\columnwidth]{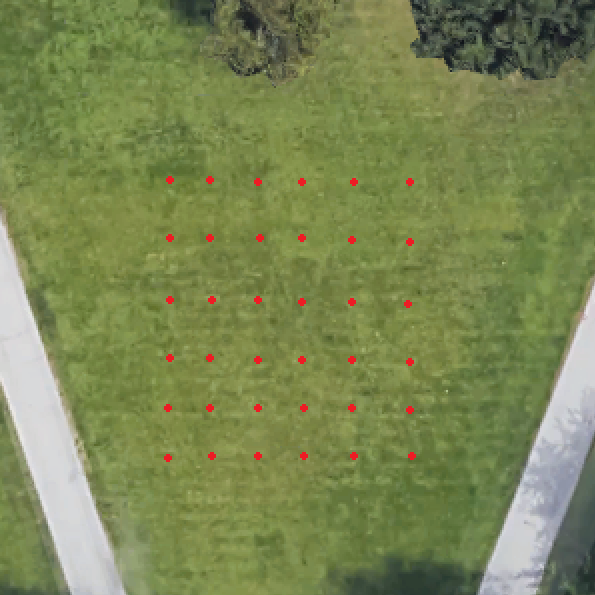}}\quad{}\subfloat[]{\includegraphics[viewport=0bp 0bp 476bp 476bp,width=0.47\columnwidth]{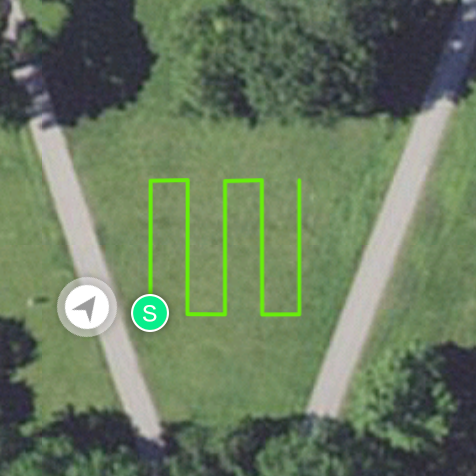}}\caption{\label{fig:Scenarios} a) Insect positions (red dots) in our experimental
setting. b) The DJI Ground Station Pro interface displaying the WayPoint
routing mission. The white and gray arrow represents the takeoff location. }
\end{figure}

Due to regulatory restrictions and local weather conflicts, opportunities
for testing the method in an outdoor field at night, are limited.
Therefore, we conducted our tests on a campus field located at Marquette
University in Milwaukee, Wisconsin. To mitigate the impact of the
illumination at the edges of the field, all the tests were performed
on a $15.24\times15.24$ meter area at the center of the field (Figures
\ref{fig:Scenarios}(a) and \ref{fig:Scenarios}(b)). In the test
area, the average grass height was 9 cm, the average luminous value
was 0.1 lux, and the average wind speed during testing was 1.4 km/h.

To make the conditions in our testing field as similar as practically
possible to the natural fields insects would inhabit, we placed some
insects on top of the grass gently and some closer to the base of
the grass, causing them to be slightly occluded from above by the
grass. The arrangement of the insects can be seen in Figure \ref{fig:Scenarios}(a).
A total of 36 insects were arranged in six rows of six insects each
with gaps of 3.05 meters between each other.  The data was collected
with the UAV flying at a height of 10 meters. The resolution of the
video was $4096\times2160$. Under these conditions, the average width
of the insects was 7.8 pixels and the average length was 14.6 pixels.
The standard deviation of these values were 2.7 pixels in width and 5.0 pixels in length. The UAV flying speed was 1m/s and the video frame rate
was 24 frames per second. Once the system has been set up, the entire
data collection session can be performed in approximately three minutes.

\subsection{Mission planning}

To ensure that all flight tests are performed in a similar pattern
with minimal human error, we use the DJI Ground Station (GS) Pro app\footnote{https://www.dji.com/ground-station-pro}
to prepare a flight pattern ahead of the experiments. The tool provides
planning flexibility, allowing many parameters to be adjusted, such
as flight speed, flight altitude, and corner rounding radius. In this
project, we used the WayPoint Route mission type (Figure \ref{fig:Scenarios}(b)),
which provides a simple tap-to-mark system to select the locations
to which the UAV should fly. Based on the user-provided locations,
the application creates a set of GPS waypoints that intercept these
points while also satisfying the image acquisition requirements. As
Figure \ref{fig:Scenarios}(b) shows, in our experiments, we used
10 waypoints with a spacing of 15.24 meters in the vertical (i.e., south
to north) direction and 3.05 meters in the horizontal (west to east) direction. 

\subsection{Insect detection performance}

\begin{figure}[tbh]
\centering{}\includegraphics[width=0.97\columnwidth]{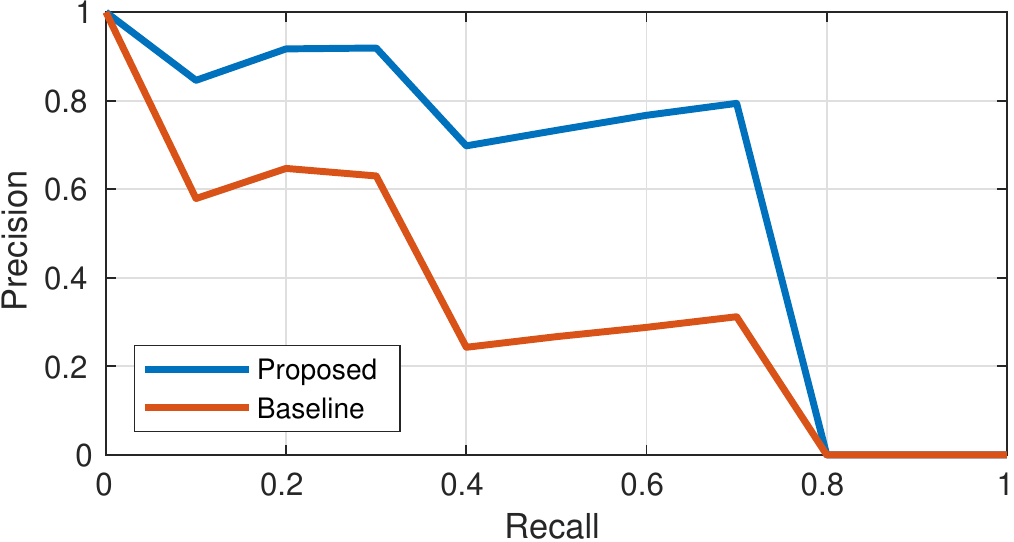}\caption{\label{fig:Precision-recall-curves}Precision and recall curves for
the baseline and proposed insect detection algorithms generated using
the Multiple Object Tracking Development kit \cite{MOTChallenge2015}.}
\end{figure}

In the evaluation of the detection performance, we chose not to rely
on the GPS coordinates of the targets or the UAV as the ground truth.
This approach eliminates the effects of localization and georeferencing
error in our analysis. Placing insects in a formation with specific
GPS coordinates is prone to error, especially with low precision methods
to determine the GPS location of a specific spot in the field. Instead,
we evaluate the insect detection method directly on the images by
comparing the location and size of the bounding boxes around each
detection $d_{i}$ in $\mathbb{I}_{i}^{(r)}$ with the ground truth
boxes ($GT_{i}$). $GT_{i}$ is a $h'\times w'$ binary image that
contains manually labeled detection targets. To be labeled as a ground
truth target, an insect must be clearly visible to the naked eye within
the UV light beam in the images. A detection is considered a true
positive if its intersection over union (IoU) metric with respect
to a ground truth bounding box is higher than 50\%. We compute the
average recall and precision values of the baseline and proposed methods
over the entire video sequence \cite{Fawcett_2006_ROC}. In our experiments,
the testing footage consists of 2477 frames. Figure \ref{fig:Precision-recall-curves}
shows the the precision-recall (PR) curves for our approach and the
baseline method. The area under the PR curve for our method is $0.61$
whereas for the baseline approach it is $0.36$. Our method shows
a precision as high as $0.8$ for a recall rate of $0.7$, while the
baseline methods has a precision of $0.28$ for the same recall value.
Most of the mistakes in our method are due to the fact that an IoU
of 50\% is challenging to achieve for very small objects. At lower
IoU levels, the results would be improved.

\subsection{Computation time}

The uncomplicated design of the detection algorithm allows us for
quick processing of footage. The image cropping step can be executed
in $O(1)$ time, and the remaining steps run in $O(s')$ time, where
$\mathit{s'=h'\,\mathrm{x}\,w'}$, i.e., the resolution of the region
of interest. The overall method runs in $O(s')$ time.

\section{Conclusions and Future Work}

We have described a novel system that combines UAVs, ultraviolet lighting
systems, and computer vision algorithms to detect fluorescent-coated
insects in the field. To the best of our knowledge, this is the first
vision-based system that detects insects in the field. It uses an illumination system based on a UV light source to visualize the insects and a color-based detection algorithm that requires minimal calibration. The system was evaluated relative to its detection precision and recall and compared with a baseline detection approach based on Otsu\textquoteright s
algorithm. The proposed system corresponds to a significant advancement
of the state of the art as manual insect recapture rates are much
lower, even with long range laser-based systems \cite{Rice_2015_Laser}. 

In the future, we intend to incorporate the GPS locations of the insects
as the ground truth labels and use the UAV's GPS location and inertial
measurements to generate orthographic projections of the insect locations.
Being able to use GPS locations rather than labeled video footage
would allow us to directly map the locations of invasive insects and
monitor their dispersal.

\section*{Acknowledgment}

This work is supported by Agriculture and Food Research Initiative Agricultural Engineering grant no. 2018-67021-28318 from the USDA National Institute of Food and Agriculture. Any opinions, findings, conclusions, or recommendations expressed in this publication are those of the author(s) and do not necessarily reflect the view of the U.S. Department of Agriculture.
 
The authors would like to thank Weihua Liu and Scott Stewart for their
assistance with data collection and annotation.


{\small
\bibliographystyle{ieee}
\bibliography{Bug_Paper_new}
}

\end{document}